\documentclass{edm_article}

\usepackage[T1]{fontenc}
\usepackage{graphicx}
\usepackage{amsmath}
\usepackage[colorlinks=true, allcolors=blue]{hyperref}
\usepackage{multirow}
\usepackage{array}
\usepackage{times}
\usepackage{amsmath}
\usepackage{algorithm}
\usepackage{algorithmic}
\usepackage{wrapfig}
\usepackage{multirow}
\usepackage{graphicx} 
\usepackage{makecell}  
\usepackage{adjustbox}
\usepackage{placeins}

\title{Just Read the Question: Enabling Generalization to New Assessment Items with Text Awareness}

\numberofauthors{4}
\author{
 \alignauthor
Arisha Khan\textsuperscript{*}\\
       \affaddr{Carleton College}\\
       \email{khana2@carleton.edu}
 \alignauthor
Nathaniel Li\textsuperscript{*}\\
       \affaddr{Carleton College}\\
       \email{lin@carleton.edu}
 \alignauthor
Tori Shen\textsuperscript{*}\\
       \affaddr{Carleton College}\\
       \email{shent@carleton.edu}
       \and
 \alignauthor
Anna N. Rafferty\\
       \affaddr{Carleton College}\\
       \email{arafferty@carleton.edu}
}

\date{}

\begin{document}
\maketitle
\footnotetext[1]{* Equal contribution, listed in alphabetical order.}

\begin{abstract}
Machine learning has been proposed as a way to improve educational assessment by making fine-grained predictions about student performance and learning relationships between items. One challenge with many machine learning approaches is incorporating new items, as these approaches rely heavily on historical data. We develop Text-LENS by extending the LENS partial variational auto-encoder for educational assessment to leverage item text embeddings, and explore the impact on predictive performance and generalization to previously unseen items. We examine performance on two datasets: Eedi, a publicly available dataset that includes item content, and LLM-Sim, a novel dataset with test items produced by an LLM.
We find that Text-LENS matches LENS' performance on seen items and improves upon it in a variety of conditions involving unseen items; it effectively learns student proficiency from and makes predictions about student performance on new items.
\end{abstract}


\section{Introduction}

Standardized assessments are commonly used in schools, with legal mandates for testing in some countries such as the US~\cite{nea2025essa}. Machine learning approaches to student modeling have been proposed as a way to more efficiently use test data and better align reporting from assessments with the needs of teachers and students (e.g.,~\cite{christie2023lens,wang2020varfa,wu2020variational,zhai2020applying}). These approaches typically require a large amount of data about student performance on each test item. This poses a challenge: the introduction of new test items. Inferences about new items cannot be made until significant student performance data about these items are available.  This  ``cold start'' problem, which also exists for typical psychometric approaches, leads to longer assessments in order to field test items~\cite{haladyna2013developing}. Further, existing models must be retrained to incorporate new items, a resource- and time-intensive process.

To address this challenge, we modify an existing ML model for educational assessment, LENS~\cite{christie2023lens}, to leverage text embeddings from a transformer encoder. This model, Text-LENS, can learn more expressive, text-aware item representations.

We also leverage a large language model (LLM) for synthetic test item generation to examine our method's performance in a controlled environment.

We experimentally investigate the effectiveness of Text-LENS and show strong performance when making predictions about performance on seen items based on seen or unseen items. 
Text-LENS also makes informative predictions about previously unseen items based on the same input information, although performance is not as strong as when the query item is seen. Performance on the simulated dataset demonstrates that Text-LENS incorporates information about both item difficulty and student proficiency, and that it implicitly learns about relationships among items, such as whether they target the same skills. Overall, these results show the potential of this approach as a scalable and effective alternative to assessment models that rely solely on item identity as opposed to content.

 \section{Related work}\label{sec:related}

Leveraging text embeddings is increasingly common for educational tasks, including short answer grading~\cite{putnikovic2023embeddings} and alignment of educational standards~\cite{butterfuss2024application}. Embeddings of item text have been used to assess properties of assessment items: for instance, to identify similar questions in a large item pool~\cite{liu2018finding} or to classify items based on Bloom's taxonomy~\cite{gani2023bloom}. Most closely related to the properties we seek to learn about, several papers have used embeddings to estimate item difficulty (see~\cite{benedetto2023survey} for a survey). For example, neural networks have been trained on embeddings of the item text and related materials (e.g., a reading passage for a reading comprehension question)~\cite{huang2017question,lin2019automated}, and multiple choice item difficulty have been estimated by comparing item and answer choice embeddings~\cite{hsu2018automated}.
In our work, we explore whether a partial VAE-based architecture can learn about item difficulty and relations among items based on text in order to accurately predict student performance.

Several papers have proposed using question text as part of knowledge tracing. For example, one proposal used LSTMs and attention to learn from both embeddings of item text and historical performance data~\cite{su2018exercise}, finding good performance given a sufficiently large training set. 
Relation-aware knowledge tracing explicitly learns relations among items from embeddings and performance data~\cite{pandey2020rkt}, outperforming~\cite{su2018exercise}.  In the context of psychometric approaches like IRT, several recent papers attempt to address the problem of new items through text embedding, either with or without a small amount of preliminary student data on these items~\cite{ma2024diffusion,mccarthy2021jump}.

One challenge of exploring these approaches is the lack of publicly available data. Many papers, such as~\cite{su2018exercise}, use proprietary data. Pandey et al.~\cite{pandey2020rkt} scraped several public databases to extract item text; we appreciate that they have made these datasets available, but note quality concerns: many item texts are listed as ``TIMEOUT\_ISSUE'' or the text is insufficient for a human to answer the question (e.g., ``Even throwing a dice twice, there are points and 5 5 is the number of chances?''). In assessment-specific work, Benedetto et al.~\cite{benedetto2023survey} point to item confidentiality and lack of public availability as key barriers to research involving item text. Here, we explore text embeddings for predicting performance on assessments, rather than knowledge tracing tasks, and we use LLM-generated item text to establish proof-of-concept results and to examine the consequences of data structure on model generalization.

 \section{Learning from question text}

We introduce Text-LENS, a modification of the LENS model that incorporates question text -- instead of learning representations based solely on question identifiers, Text-LENS uses embeddings from a pre-trained transformer encoder.

The base LENS model uses a partial variational auto-encoder (VAE) architecture to produce probabilistic representations of students based on assessment performance~\cite{christie2023lens}. LENS takes two inputs: a set of input items and a set of query items for prediction. Input items include the student correctness on the item. All items are identified to the model by their item id, which is passed through a single feed forward layer to produce an item embedding.

In LENS, the encoder concatenates the student responses and embedded item ids before mapping to the latent space. To make predictions, samples are taken from the latent space, concatenated with the embedded query item ids for prediction, and then decoded to make predictions for query item responses. In Text-LENS, both of the item id embeddings are replaced by text embeddings of the content of these items.

LENS' item id-based embeddings represent relationships between known items but do not encode item content and meaningful interpolations cannot be made between them. In contrast, text-based item embeddings map into continuously meaningful space, even when the particular text of an item has never been seen by the model before. The text-based embeddings also have the potential to encode the difficulty of the items without further training. These qualities improve the generalization of item embeddings and reduce the need for retraining for good performance.

We expect that text embeddings can address several challenges that arise in educational assessment settings. First, text embeddings will allow the model to learn from new items (not present in training) by mapping these new items into the same semantic space. Second, text-awareness will allow the model to align input items with the query item and infer the difficulties of items for better prediction on unseen query items.

 \section{Datasets}\label{synthetic-data}

Testing the effectiveness of Text-LENS requires datasets where item text and student performance data are available. There are limited publicly available datasets that meet these requirements. We thus use both synthetic and real-world data: the synthetic dataset allows for more fine-grained control and checking or assumptions, while the real-world datasets validate these results in a realistic setting.

\textsc{Eedi}: We used data from   Task 3 \& 4 of the NeurIPS 2020 Education Challenge~\cite{wang2020diagnostic}. It consists of responses to multiple choice math items from two school years on the Eedi platform. 
Item content is provided as images, which we processed using Optical Character Recognition (OCR). OCR introduced considerable noise: Mathematical formulas are not always accurate, and sequencing of item components was occasionally distorted in the output text.

We filtered the data to include only five distinct skills from the skill taxonomy in order to facilitate experimentation with the relationship between input and query items. After randomly selecting among those skills associated with at least 20 items, the data included $4,641$ students and $214$ unique items.

\textsc{LLM-Sim}: This fully synthetic dataset consists of items generated by prompting GPT-4o about specific topics, eliminating the OCR artifacts while focusing on similar topics. The LLM was told to generate ten easy, medium, and hard math test questions for the given subject, and asked to ``significantly vary the wording of the questions, such as the type of information included in the question and the phrasing of the task itself.'' Fourteen topics were randomly selected from level 3 (of 4 total levels) of the Eedi skill taxonomy. For each skill, we prompted the LLM separately about each of its 3-4 subskills. This resulted in 90-120 items per skill.\footnote{Full dataset available \href{https://osf.io/semd4/?view_only=a7a6846e20f945a186de79a786a3c0ff}{here}.} Since later experiments needed only five skills and 40 items per skill,  we filtered the data to randomly select only the needed number of items, resulting in a total of 200 items (40 items for each of 5 skills).

After generating item text, we created $50,000$ simulated students and sampled their responses to all items via an IRT model. For each simulated student, we independently sampled one proficiency parameter per skill; proficiency parameters were sampled from~$\mathcal{N}(0,1)$, clipped to fall in $[-4,4]$. By structuring the five skill proficiencies to be independent of one another, we can investigate whether Text-LENS correctly learns relationships between items and skills.
Items were assigned difficulty parameters of $-1.5$ (easy), $0$ (medium), or $1.5$ (hard) based on the LLM's estimated difficulty level when generating the question. 
For each item, the probability the student would answer correctly was calculated according to a 3-PL IRT model~\cite{van2016handbook} with discrimination parameter $1$ and guessing parameter $0.1$.
To enable a variety of experimental structures, we create simulated responses for all 200 items for every student.

 \FloatBarrier 
\begin{figure*}
    \Description[Results for LENS and Text-LENS on LLM-Sim dataset.]{Text-LENS outperforms LENS on both on-target and off-target queries when the input is seen or unseen.}
    \includegraphics[width=.45\textwidth]{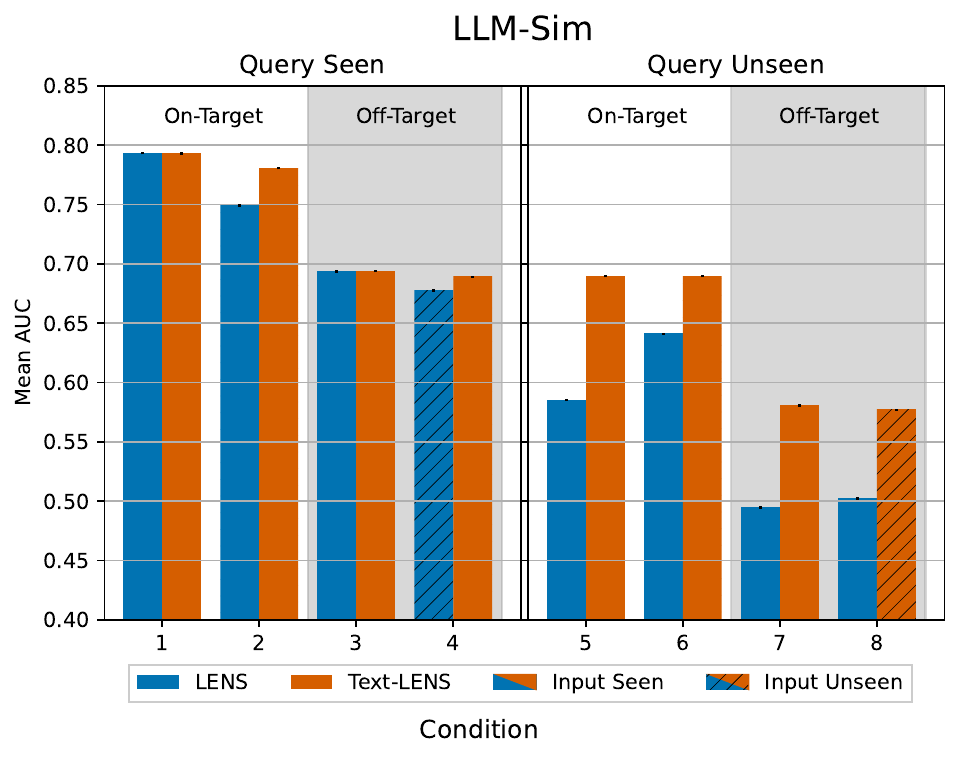}
    \hfill
    \Description[Results for LENS and Text-LENS on Eedi dataset]{Text-LENS outperforms LENS on both on-target and off-target queries when the input is seen or unseen. Effects are less pronounced on Eedi than on the LLM-Sim dataset. In Eedi, dependencies among skills mean that off-target input is still relevant for estimating student proficiency on query item.}
        \includegraphics[width=.45\textwidth]{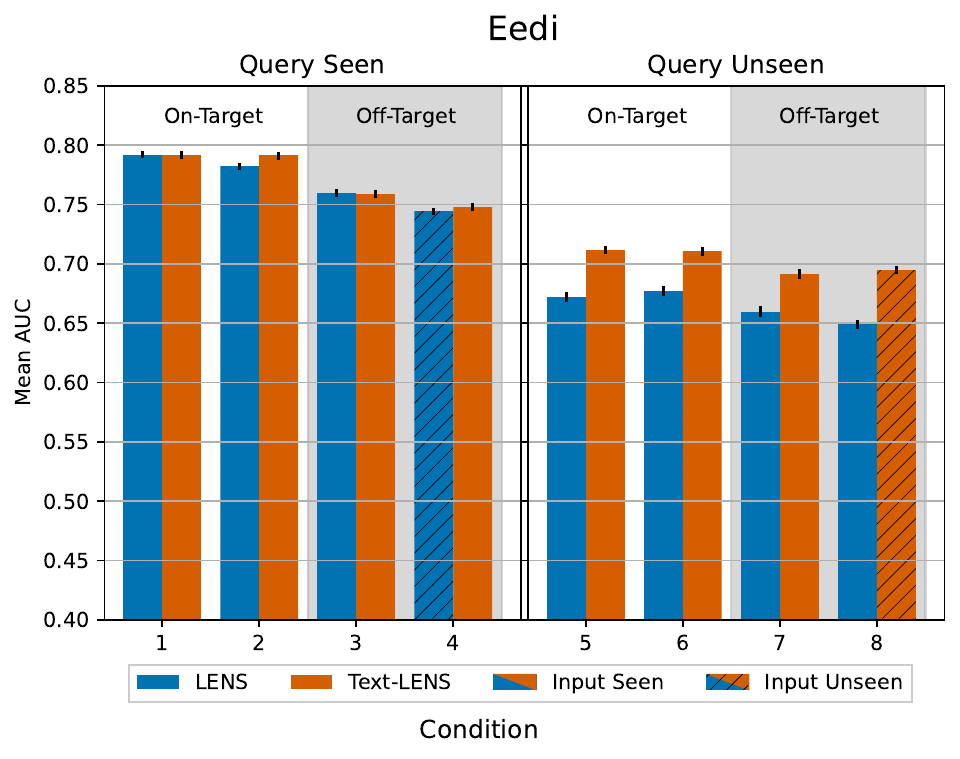}
    \caption{LENS and Text-LENS performance on LLM-Sim (left) and Eedi (right) across all conditions. The error bars represent one standard error over 60  repetitions.}\label{fig:resultgraphs}
\end{figure*}

\section{Experiments}

Our experiments aim to assess how effectively Text-LENS leverages item text to improve assessment predictions. We test Text-LENS' ability to extract item information and understand the degree to which it can make generalizations about new (unseen) items. Specifically, we investigated the following research questions:

\begin{itemize}
\item \textbf{Performance on seen queries:} How well does Text-LENS incorporate information from seen or unseen input to make predictions about performance on seen query items?
    \item \textbf{Performance on unseen queries:} How effectively can Text-LENS make predictions about student performance on unseen query items?
    \item \textbf{Varying relationships between input and query:} How does the relationship between the input items and the query items affect performance predictions?
\end{itemize}

\subsection{Training and Item Embeddings}
We evaluate LENS and Text-LENS on Eedi and LLM-Sim. We train on 80\% of students and use 10\% each for validation and testing.
We partitioned the items into halves. Models were trained on one half, reserving the remaining items as unseen.

For these experiments, we used mathBERT as our pretrained transformer encoder to produce the item embeddings for Text-LENS \cite{devlin2019bert}. The original LENS model serves as our performance baseline.

We trained LENS and Text-LENS on each dataset using the model parameters from the original LENS paper, but performed grid search on learning rate (LR), distribution dimensions, encoder hidden dimensions, and accumulator hidden dimensions for each of our datasets and for both LENS and Text-LENS. These parameters were selected for their relevance to the model's ability to represent the number of skills in a dataset. All models were trained using the Adam optimizer for 300 epochs.

\begin{table}[h]
\centering
\caption{Parameters used in training of each model.}
\small
\renewcommand{\arraystretch}{1.2}
\setlength{\tabcolsep}{4pt} \begin{tabular}{|@{\hspace{2pt}}c@{\hspace{2pt}}|c|c|c|c|c|}
\hline
\textbf{Dataset} & \textbf{Model} & \textbf{LR} & \textbf{\makecell{Dist. \\ Dim.}} & \textbf{\makecell{Encoder \\ Hidden Dim.}} & \textbf{\makecell{Accumulator \\ Hidden Dim.}} \\
\hline
\multirow{2}{*}{Eedi} 
    & LENS      & .001  & 16  & 30  & 8  \\ \cline{2-6}
    & Text-LENS & .001  & 16  & 30  & 8  \\
\hline
\multirow{2}{*}{LLM-Sim} 
    & LENS      & .001  & 64  & 90  & 24 \\ \cline{2-6}
    & Text-LENS & .005  & 64  & 30  & 24 \\
\hline
\end{tabular}

\end{table}

\subsection{Test Setup}

We evaluated our models to assess performance with unseen items. For each test student, the model received a query item to predict and a set of input items containing student responses. The skill associated with the query item was designated as the target skill. If the input items were selected to be of the same skill as the query, we consider them to be on-target.

\begin{table}[h]
\centering
\caption{Experimental conditions varying which items are previously seen and the relationship between input and query items.}
\begin{tabular}{|c|c|c|c|c|c|}
\hline
\multirow{2}{*}{\textbf{Conditions}} & \multicolumn{4}{c|}{\textbf{Input Items}} & \multirow{2}{*}{\textbf{Query}} \\ \cline{2-5}
 & \multicolumn{2}{c|}{\textbf{Target Skills}} & \multicolumn{2}{c|}{\textbf{Other Skills}} &  \\ \cline{2-5}
 & \textbf{Seen} & \textbf{Unseen} & \textbf{Seen} & \textbf{Unseen} &  \\ \hline
 
1 & 19 & 0 & 0 & 0 & \multirow{4}{*}{Seen} \\ \cline{1-5}
2 & 0 & 19 & 0 & 0 &  \\ \cline{1-5}
3 & 0 & 0 & 19 & 0 &  \\ \cline{1-5}
4 & 0 & 0 & 0 & 19 &  \\ \hline
5 & 19 & 0 & 0 & 0 & \multirow{4}{*}{Unseen} \\ \cline{1-5}
6 & 0 & 19 & 0 & 0 &  \\ \cline{1-5}
7 & 0 & 0 & 19 & 0 &  \\ \cline{1-5}
8 & 0 & 0 & 0 & 19 &  \\ \hline
\end{tabular}
\label{tab:experimental_conditions}
\end{table}

As shown in Table~\ref{tab:experimental_conditions}, our experimental design varied three factors: (1) whether the input items were seen or unseen during training, (2) whether the query item was seen or unseen during training, and (3) whether the input items' skill was on- or off-target. These conditions allows us to assess how effectively text embeddings enable generalization to novel questions compared to ID-based representations across both synthetic and real-world datasets.

\section{Results}

We find that Text-LENS consistently matches or surpasses LENS, highlighting the potential of text-awareness to enhance test inference flexibility and performance.

\subsection{Performance on seen queries}

When all input and query items have previously been seen in training, Text-LENS and LENS show very similar performance (Figure~\ref{fig:resultgraphs}, conditions 1 and 3). 
This parity indicates that text-based embeddings can effectively capture at least as much information as ID-based embeddings for items that have been seen during training.

Text-LENS uses the embeddings of unseen input items to effectively make inferences about previously seen query items: it performs as well (Eedi) and nearly as well (LLM-Sim) in condition 2 as in condition 1 (Figure~\ref{fig:resultgraphs}). Meanwhile, LENS' performance in condition 2 is somewhat degraded.
Both models still extract meaningful information from unseen input items since the items still generally characterize the student’s skill. This insight explains LENS’ strong performance even when the input is unseen on condition 2: higher performance on the input is strongly correlated with success on the query item, mirroring the pattern from training.

\subsection{Performance on unseen queries}

In all conditions with unseen queries, Text-LENS achieves a higher AUC than LENS (right panels in Figure~\ref{fig:resultgraphs}). 
However, inferences about unseen query items are less accurate than inferences about seen query items: for instance, when input items are seen but query unseen (condition 5), Text-LENS' AUC drops $0.08$ on Eedi and $0.10$ on LLM-Sim compared to when query items were seen (condition 1). LENS falls more sharply, decreasing $0.12$ on Eedi and $0.20$ on LLM-Sim.
Unseen queries are associated with a larger drop in performance than unseen input (condition 1 vs.~5 compared to condition 1 vs.~2). 
This larger penalty likely reflects the fact that with unseen input, the model receives both an item identifier and student performance on that item, but only the identifier is available for the query.

\subsection{Varying input relevancy}

In the previous comparisons, the target skill of the input items aligned with that of the query item. The models' inferences depend on both estimates of student proficiency on this skill and estimates of the difficulty of the query item. When the input items are aligned with a different skill than the query item, the model must fall back to using estimates of query item difficulty and any learned relationships among skills. In LLM-Sim, skills are independent, whereas relationships among skills may exist in Eedi.

Off-target query items consistently result in lower performance compared to the analogous on-target query conditions (see Figure~\ref{fig:resultgraphs}), especially for LLM-Sim. Text-LENS seems to learn about query item difficulty: for conditions 7 and 8, LENS performs no better than random, but Text-LENS achieves an AUC of about 0.58. Since the input items are off-target and the query item is unseen in training, Text-LENS could only have made its inference by extracting difficulty from text, and its better-than-random performance indicates that it is doing so. To further confirm, we simulated and retrained the models on LLM-Sim where the relationship between item text and difficulty was randomized and observed the Text-LENS' performance drop to chance.

While there is still some performance drop for off-target query items on Eedi, this drop is much more modest. This likely reflects that, while the skill alignment between input and query is relevant to inference, there are also strong dependencies between skills that allow off-target skills to provide good information about the student’s response to the query item. 

\subsection{Performance across different datasets}

While the relative performance between Text-LENS and LENS for any particular condition is preserved across both datasets, we find some performance differences between our datasets.

Surprisingly, on LLM-Sim, LENS performs better when making inferences about unseen query items with unseen input items (condition 5) than with seen (condition 4). Here, the seen input items could be misleading, since the item ID-based embedding method can create meaningful embeddings for the input, but any relationship between the input and query embeddings is spurious. In contrast, LENS’ input item embeddings are unreliable in condition 5 -- just as in condition 2. Here, the model may infer based only on a student’s general performance and obtain good results by doing so.

This increase in LENS' performance from condition 4 to 5 is not present on Eedi. Here, due to dependencies between skills, it is not as important for the model to understand the relationships between the input and query. This also explains the smaller performance change between conditions 1 and 2 on Eedi than on LLM-Sim.

Text-LENS performs consistently across both datasets, with no unpredictable behavior like LENS' performance jump from condition 4 to 5, except for the greater impact of off-target items in LLM-Sim due to complete skill independence.

 \section{Discussion}

While traditional assessment models require extensive historical performance data for each new item, our text-aware approach allows immediate integration of unseen questions, improving the ``cold start'' problem. Text-LENS could significantly reduce the need for field testing new items before operational use, shortening assessment length and decreasing student testing time. Additionally, this approach enables dynamic adaptation of assessments—instructors can modify questions or introduce entirely new ones while maintaining predictive accuracy about student performance. Our experiments demonstrate that text embeddings effectively encode item difficulty and skill relationships directly from question content, allowing for more flexible assessment design and reducing the resource-intensive retraining typically required when item pools evolve. 

Natural directions for future work include further investigation into real-world performance and the value of LLM-based simulated data. Our evaluation on LLM-Sim indicates the ability of text embedding to encode difficulty, but it is unclear to what degree real-world questions encode difficulty in text and how that differs from an LLM specifically prompted to produce questions at different difficulties. This evidence is in alignment with some prior work (see Section~\ref{sec:related}), but that work was not focused on ML models for assessment and was typically focused on non-math subjects. A high-quality assessment dataset with test content and either professionally annotated difficulties or difficulties inferred by an IRT-based system would help to investigate the degree to which Text-LENS can recover item difficulty from text.

Our evaluation of real-world data was limited to Eedi, for which test content only exists as screenshots of items. Thus, our item embeddings were produced from noisy and sometimes incorrect OCR. Furthermore, any graphical or spatial information included in the question was lost. Future work could attempt to incorporate multimodal items; prompting a multimodal LLM to summarize the item in text might be one path forward. A further limitation is our focus solely on math problems. In Eedi, this might lead to high dependencies among skills compared to some other subjects. Math questions may rely more on diagrams and have less rich text, so Text-LENS might exhibit stronger performance on subjects where language features more prominently, such as reading comprehension.

The text of exam questions provides a rich source of information that can be applied toward response inference by leveraging text embeddings. While further evaluation is needed prior to deployment for real-world applications, the strong performance of Text-LENS across a variety of realistic use cases demonstrates the feasibility of using text embeddings for ML-based assessment models.

\newpage
\section{Acknowledgements}

We thank Aadi Akyianu for their initial work on the project, and Jared Arroyo-Ruiz, Geoffrey Jing, and Nhi Luong for prior involvement in this research project. We also thank Carson Cook and S. Thomas Christie for early discussions about the project idea and direction.
\bibliographystyle{splncs04} \bibliography{bibiliography}

\balancecolumns

\end{document}